\documentclass[lettersize,journal]{IEEEtran}
\usepackage{amsmath,amsfonts}
\usepackage{algorithmic}
\usepackage{algorithm}
\usepackage{array}
\usepackage[caption=false,font=normalsize,labelfont=sf,textfont=sf]{subfig}
\usepackage{textcomp}
\usepackage{stfloats}
\usepackage{url}
\usepackage{verbatim}
\usepackage{graphicx}
\usepackage{cite}
\usepackage{booktabs}
\usepackage{multirow}
\usepackage{orcidlink}
\begin{document}

\title{DisenTS: Disentangled Channel Evolving Pattern Modeling for Multivariate Time Series Forecasting}

\author{Zhiding Liu\textsuperscript{~\orcidlink{0000-0003-0994-473X}}, Jiqian Yang\textsuperscript{~\orcidlink{0009-0007-6421-2626}}, Qingyang Mao\textsuperscript{~\orcidlink{0000-0002-6922-856X}}, Yuze Zhao\textsuperscript{~\orcidlink{0009-0007-3542-4304}}, Mingyue Cheng\textsuperscript{~\orcidlink{0000-0001-9873-7681}}, Zhi Li\textsuperscript{~\orcidlink{0000-0002-8061-7486}}, \\ Qi Liu\textsuperscript{~\orcidlink{0000-0001-6956-5550}}~\IEEEmembership{Member,~IEEE} and Enhong Chen\textsuperscript{~\orcidlink{0000-0002-4835-4102}}~\IEEEmembership{Fellow,~IEEE}
\thanks{This research was supported by grants from the Joint Research Project of the Science and Technology Innovation Community in Yangtze River Delta (No. 2023CSJZN0200), and the Fundamental Research Funds for the Central Universities. This work also thanked to the support of funding MAI2022C007. The corresponding author is Enhong Chen.}
\thanks{
Zhiding Liu, Jiqian Yang, Qingyang Mao, Yuze Zhao, Mingyue Cheng, Qi Liu, and Enchong Chen are affiliated with the State Key Laboratory of Cognitive Intelligence, University of Science and Technology of China, Hefei 230026, China. Email: \{zhiding, yangjq, maoqy0503, yuzezhao\}@mail.ustc.edu.cn, \{mycheng, qiliuql, cheneh\}@ustc.edu.cn.
}
\thanks{Zhi Li is affiliated with the Shenzhen International Graduate School, Tsinghua University, Shenzhen  51800, China. Email: zhilizl@sz.tsinghua.edu.cn}

}

\markboth{Journal of \LaTeX\ Class Files,~Vol.~14, No.~8, August~2021}%
{Shell \MakeLowercase{\textit{et al.}}: A Sample Article Using IEEEtran.cls for IEEE Journals}

\IEEEpubid{0000--0000/00\$00.00~\copyright~2021 IEEE}

\maketitle

\begin{abstract}

Multivariate time series forecasting plays a crucial role in various real-world applications. Significant efforts have been made to integrate advanced network architectures and training strategies that enhance the capture of temporal dependencies, thereby improving forecasting accuracy. On the other hand, mainstream approaches typically utilize a single unified model with simplistic channel-mixing embedding or cross-channel attention operations to account for the critical intricate inter-channel dependencies. Moreover, some methods even trade capacity for robust prediction based on the channel-independent assumption. Nonetheless, as time series data may display distinct evolving patterns due to the unique characteristics of each channel (including multiple strong seasonalities and trend changes), the unified modeling methods could yield suboptimal results.

To this end, we propose DisenTS, a tailored framework for modeling disentangled channel evolving patterns in general multivariate time series forecasting. The central idea of DisenTS is to model the potential diverse patterns within the multivariate time series data in a decoupled manner. Technically, the framework employs multiple distinct forecasting models, each tasked with uncovering a unique evolving pattern. To guide the learning process without supervision of pattern partition, we introduce a novel Forecaster Aware Gate (FAG) module that generates the routing signals adaptively according to both the forecasters’ states and input series’ characteristics. The forecasters’ states are derived from the Linear Weight Approximation (LWA) strategy, which quantizes the complex deep neural networks into compact matrices. Additionally, the Similarity Constraint (SC) is further proposed to guide each model to specialize in an underlying pattern by minimizing the mutual information between the representations. We conduct extensive experiments on a wide range of forecasting settings with diverse state-of-the-art forecasting models, and the results validate the effectiveness and generalizability of the proposed DisenTS framework.

\end{abstract}


\section{Introduction}

\IEEEPARstart{M}{ultivariate} time series is a ubiquitous data format consisting of multiple signals collected in chronological order, presenting real-world observations. The analysis of time series, especially time series forecasting, plays a crucial role in aiding decision-making and planning across a range of daily applications, including cloud resource allocation \cite{chen2021graph}, traffic load analysis \cite{yin2016forecasting}, financial risk assessment \cite{hou2022multi}, and climate forecasting \cite{nature, nature2}. Extensive efforts have been dedicated to the field of multivariate time series forecasting. Notably, deep learning-based approaches have demonstrated exceptional proficiency in capturing the evolving patterns within time series data \cite{nbeats, informer, quatformer, wang2024deep}.

\IEEEpubidadjcol

In the literature, the majority of existing forecasting methods typically target the forecasting aspect. These approaches tend to concentrate solely on modeling the intricate temporal dependencies inherent in the original time series data \cite{wu2021autoformer, SCINet}, with the goal of mitigating the impact of non-stationarity for stable forecasting \cite{RevIN, NST, san}, and achieving higher computational efficiency \cite{logtrans, liu2021pyraformer}. However, the inter-channel dependencies are also crucial for multivariate forecasting and the coarse channel-mixing embedding operation may fail to capture such information, ultimately leading to suboptimal results. To this end, various efforts have been devoted to modeling the cross-dimensional dependencies effectively. Crossformer \cite{crossformer} introduces an explicit channel-wise attention module, and iTransformer \cite{liuitransformer} simply applies an inverted transformer for accurate forecasting. Conversely, various recent works trade capacity for robust prediction \cite{han2024capacity} by adopting the channel-independent assumption, treating multivariate time series as multiple univariates, and achieving state-of-the-art forecasting performance \cite{DLinear, patchtst}.

\begin{figure}[htbp]
  \centering
   \includegraphics[width=\linewidth]{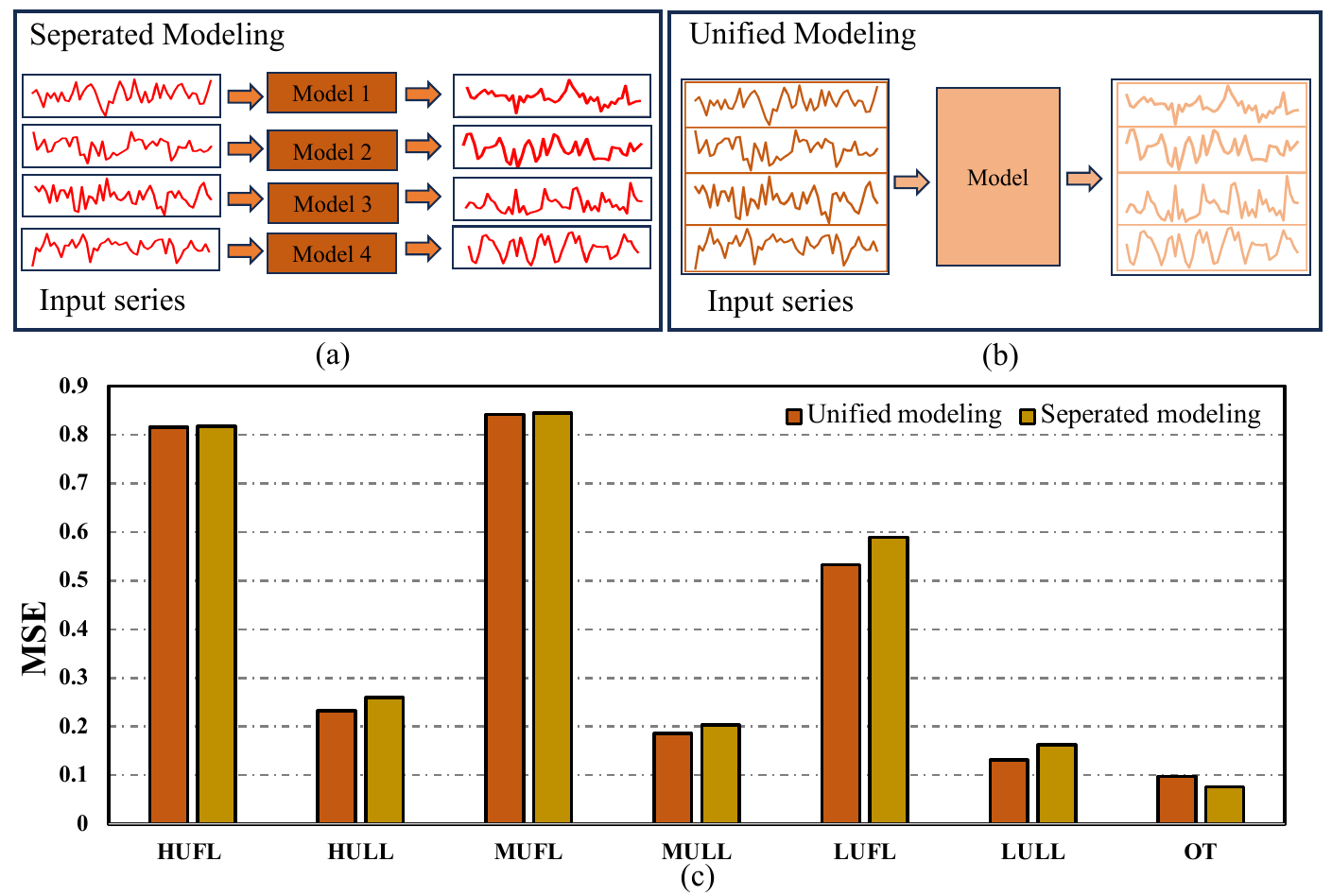}
  \caption{(a)\&(b) demonstrate the schema of the separated modeling and unified modeling. (c) provides the MSE comparison on each channel of the ETTh2 dataset with the above two schemas. The unified modeling approach drags down the performance on the Oil Temperature (OT) while promoting precision on other channels of external power load.}
  \label{fig:motivation}
\end{figure}

While these methods have proven effective, they predominantly assume homogeneity among channels and adopt a unified pattern modeling scheme where a single model is applied to all input channels, as depicted in Fig. \ref{fig:motivation}(b). Furthermore, the channel-independent-based approaches even assume a certain level of uniformity across all channels, as the neural transformation function for each channel is the same. In contrast, we posit that the collected real-world multivariate time series data may exhibit distinct evolving patterns, including multiple strong seasonalities and trend changes, due to the heterogeneity of each channel, which current methods struggle to address. For example, depending on the functionality of the city area where the sensor is located, the collected traffic flow data will show different flow patterns \cite{wang2022st}. To substantiate our argument, we conducted a preliminary experiment comparing the performance of traditional unified modeling (multivariate predict multivariate) and separated modeling (univariate predict univariate) as shown in Fig. \ref{fig:motivation}(c). We report each channel's Mean Squared Error (MSE) evaluation on the ETTh2 \cite{informer} dataset, where the OT channel represents oil temperature and the others denote external power load. Clearly, the unified modeling approach leads to worse performance on the OT channel, while yielding better results on the others. This outcome suggests the presence of multiple heterogeneous trends within the dataset and indicates a similar pattern shared among the six external power load channels. Consequently, the complex and diverse evolving patterns between channels call for a more nuanced approach.

Intuitively, a practical solution is to decouple the original time series into explicitly different components such as trend, seasonality, and holidays \cite{decompose,taylor2018forecasting}, and utilize multiple models to capture the potentially diverse evolving patterns \cite{wu2021autoformer, DLinear}. However, in real-world scenarios, time series often exhibit complex patterns that are difficult to explicitly decompose. Therefore, we introduce an easy-to-use framework tailored for modeling multiple channel evolving patterns in multivariate time series forecasting in an implicitly disentangled manner, named \textbf{\textit{DisenTS}}. Taking inspiration from the concept of a mixture-of-experts \cite{moe}, DisenTS comprises multiple experts, which can be arbitrary forecasting models, with each model assigned to uncovering a unique evolving pattern. Consequently, the framework is well-equipped to effectively address the impact of heterogeneous patterns by employing distinct backbone models. To guide the learning process without supervision of pattern partition, we introduce a novel Forecaster Aware Gate (FAG) module that generates the routing signals adaptively according to both the forecasters’ states and input series’ characteristics. Additionally, to feasibly quantify the forecasters' intricate states, we propose a Linear Weight Approximation (LWA) strategy. This method represents the complex neural forecasters using compact matrices in a channel-independent manner. Furthermore, to ensure that each forecaster specializes in a distinct underlying pattern, we incorporate a Similarity Constraint (SC) that minimizes the mutual information between intermediate representations, preventing the backbones from converging into the same representation space. Extensive experiments on a wide range of real-world datasets demonstrate that the proposed DisenTS framework can be seamlessly integrated into mainstream forecasting models, significantly enhancing their performance across various settings.

In summary, our main contributions are as follows:
\begin{itemize}
    \item We propose DisenTS, a model-agnostic framework that incorporates multiple distinct forecasting models in a mixture-of-expert manner, effectively modeling the disentangled channel evolving patterns for precise multivariate time series forecasting.
    \item We design a novel Linear Weight Approximation approach that quantifies the intricate transformation functions of neural forecasting models with compact matrices, guiding the generating of the routing signals with Forecaster Aware Gate. Moreover, explicit disentanglement on each forecasting backbone is realized through the Similarity Constraint on the matrix representation space.
    \item We conduct comprehensive experiments on a wide range of real-world time series datasets, instantiating our proposed DisenTS framework with state-of-the-art channel-independent and even channel-dependent models. The results unequivocally demonstrate the effectiveness and generalizability of our approach, as DisenTS consistently boosts the performance under different settings.
\end{itemize}

\section{Related Works}
\subsection{Time Series Forecasting}
Time series forecasting has been a prominent area of research in recent decades. Most existing studies focus on modeling temporal dependencies in time series data within a sequence-to-sequence paradigm. Early approaches, such as statistical methods like ARIMA \cite{box1968some, zhang2003time}, rely on constructing autoregressive models and forecasting through moving averages. However, these methods often require data to exhibit ideal properties, which may not align with real-world complexities. With the rise of deep learning, numerous neural forecasting models have since emerged. Recurrent Neural Networks (RNNs) were among the first architectures proposed for forecasting \cite{wen2017multi,petnehazi2019recurrent,deepar}. Despite their early promise, RNNs suffer from inherent limitations, such as a limited receptive field and error accumulation over time \cite{informer}, spurring the development of advanced architectures designed to capture long-range dependencies. Temporal convolutional networks \cite{lstnet, SCINet, moderntcn} and attention-based methods \cite{tpami_trm, wang2024diffusion} have proven particularly effective in this regard. In recent years, MLP-based methods have shown superior performance and efficiency compared to more complex neural networks \cite{DLinear,nbeats,sparsetsf}. In addition to advancements in model architecture, several specialized techniques grounded in time series analysis have been developed. These include trend-seasonal decomposition \cite{wu2021autoformer}, time-frequency transformations \cite{fedformer}, series stabilization techniques \cite{RevIN}, and patch-based approaches \cite{patchtst}.

Furthermore, self-supervised pretraining has emerged as a key research focus due to its label efficiency and generalization capabilities \cite{cheng2024learning,tpami_ssl}. Some works adopt a discriminative modeling paradigm, leveraging contrastive learning techniques to derive meaningful representations from pre-defined positive and negative pairs \cite{cost, ts2vec}. Meanwhile, generative pretraining through reconstruction has also been widely explored, with universal representations learned via masked time-series modeling \cite{tst,simmtm} or autoregressive pretraining \cite{gpht}. Notably, recent efforts have introduced large-scale corpora for training massive generative forecasting models, inspired by advancements in natural language processing (NLP) \cite{gpt3}, and pioneering works in this space have demonstrated strong performance, even in zero-shot settings \cite{timer,moirai}.

\subsection{Channel-Wise Dependencies Modeling in Time Series}
In addition to inherent temporal correlations, intricate channel-wise dependencies play a pivotal role in time series analysis, particularly in forecasting tasks. Traditional approaches often rely on channel-mixing embedding operations that project multivariate information into a single embedding representation space. However, this simplistic approach is inadequate for capturing the complex inter-channel correlations required for accurate forecasting \cite{informer, wu2021autoformer}. To address this limitation, models like Crossformer \cite{crossformer} and iTransformer \cite{liuitransformer} incorporate channel-wise attention modules, explicitly modeling channel-wise dependencies and thereby achieving improved performance. In contrast to these channel-dependent methods, recent research has adopted a channel-independent paradigm \cite{DLinear, patchtst, cheng2024convtimenet}, which treats multivariate time series as a collection of independent univariate series. By enhancing model capacity for robust prediction \cite{han2024capacity}, these methods have demonstrated superior performance on benchmark datasets. Furthermore, a recent study has utilized a mixture-of-experts architecture to capture various flow patterns influenced by the functional distributions of city areas, leading to more accurate traffic predictions \cite{wang2022st}.

\subsection{Disentangled Learning in Time Series}
Disentangled learning techniques are widely utilized in representation learning, with the primary objective of providing interpretable analytics for vectors learned by neural networks \cite{tran2017disentangled, wu2023learning}. In the domain of time series modeling, significant efforts have been made to integrate disentangled learning for various purposes, including temporal causal inference \cite{yao2022temporally} and improving domain adaptation performance \cite{li2022towards}. For more specific time series tasks, CoST \cite{cost} introduces contrastive learning objectives in both the time and frequency domains to capture disentangled seasonal-trend representations, thereby enhancing forecasting accuracy. Similarly, TIDER \cite{liu2022multivariate} applies an explicit disentanglement constraint on latent embeddings related to trend, seasonality, and local bias, leading to more accurate time series imputation while also improving the interpretability of the imputation results.

\bigskip

Despite their effectiveness, existing methods typically apply a single unified model to all input channels, assuming homogeneity or even uniformity across channels. However, we argue that heterogeneity is prevalent in time-series data, and this unified modeling approach may struggle to capture the diverse and evolving patterns present. Inspired by explicit time series decomposition, we introduce DisenTS for general time series forecasting tasks, which is a model-agnostic framework that models the evolving patterns of multiple channels in an implicitly disentangled manner without requiring predefined functional distributions or corresponding ground truth data. As a result, our approach shifts the focus of disentangled learning from disentangling the representation of the time series data to the \textit{disentanglement of the transformation function or representation within each model for capturing the potential diverse evolving patterns}, further distinguishing it from existing disentangled learning techniques.

\begin{figure*}[htbp]
  \centering
  \includegraphics[width=\linewidth]{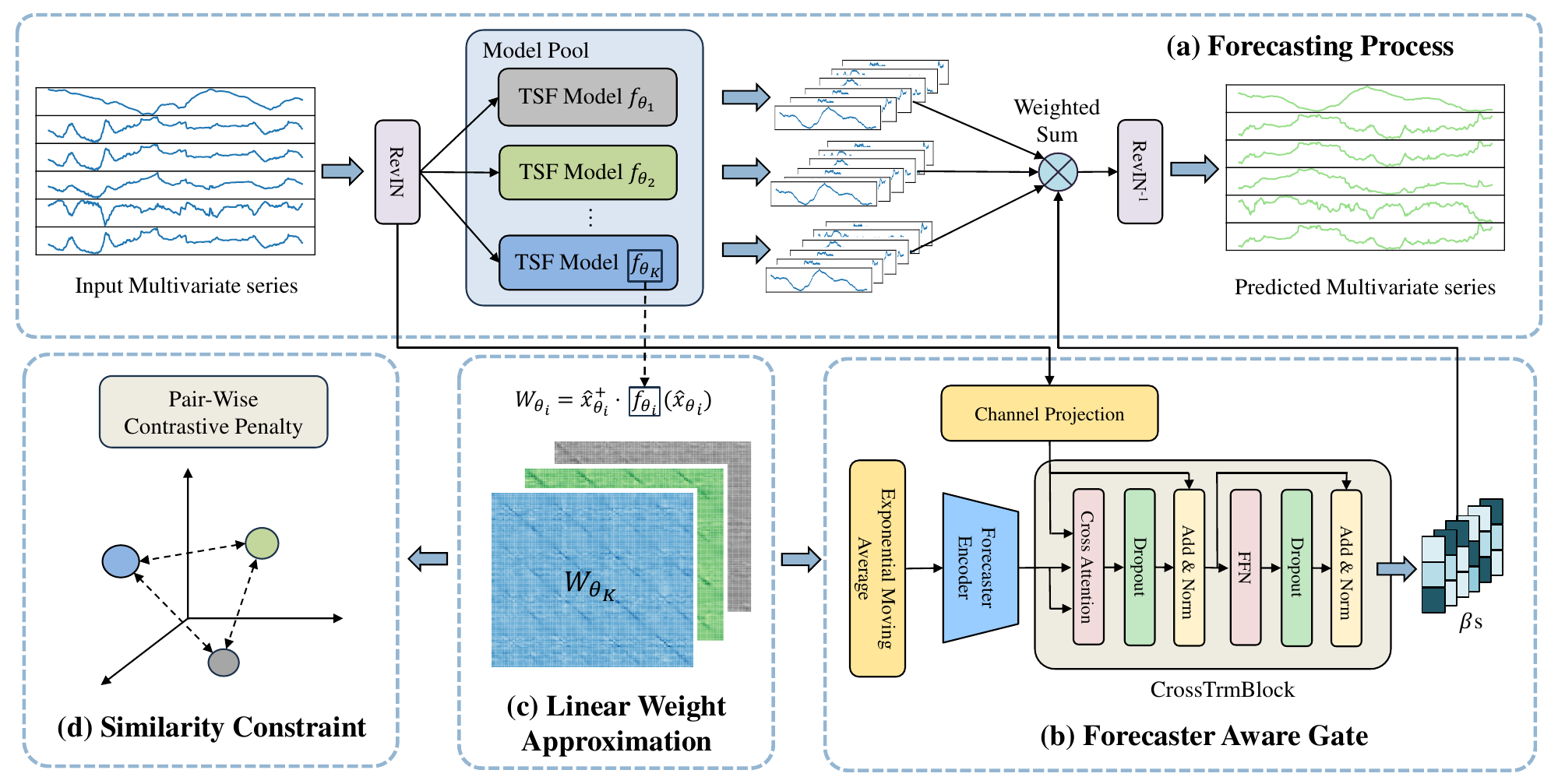}
  \caption{(a) During the forecasting stage, the framework utilizes multiple backbone models to uncover distinct evolving patterns, and the final results are predicted through a weighted sum approach. (b) The Forecaster Aware Gate (FAG) module combines the states of forecasting models and the characteristics of the input series to generate proper routing signals $\beta$ properly. (c) The Linear Weight Approximation (LWA) strategy quantifies each backbone model with a small matrix. (d) The Similarity Constraint is applied to the approximated matrices to ensure the disentanglement among the models, enforced through a pair-wise orthogonal penalty.}
  \label{fig:model}
\end{figure*}

\section{Methodology}
In this section, we provide an overview of DisenTS, a conceptually simple yet highly effective framework designed to capture the diverse evolving patterns across channels for multivariate time series forecasting. The subsequent sections will describe the overall forecasting process and introduce the three key components of the proposed framework.

\subsection{Problem Definition}

We consider a multivariate time series dataset $D \in \mathbb{R}^{C\times T}$, where $C$ denotes the number of channels and $T$ represents the total length of the dataset. The forecasting task is framed through a sliding window operation that constructs the training, validation, and testing sets. Given a set of input series $X=\{x_i\}_{i=1}^N$, the objective is to accurately predict the future values $Y=\{y_i\}_{i=1}^N$, where $x_i \in \mathbb{R}^{C\times L_{in}}$ represents the $i$-th input time series and $y_i \in \mathbb{R}^{C\times L_{out}}$ represents the corresponding target series. Here, $N$ is the number of sequences, while $L_{in}$ and $L_{out}$ denote the lengths of the input and target series, respectively.

\subsection{Forecasting Process}
As illustrated in Fig. \ref{fig:model}, the forecasting process of DisenTS can be described as a \textit{mixture of forecasting models}. It comprises a model pool with $K$ diverse forecasting models, ${f_{\theta_i}(\cdot)}_{i=1}^K$, each specialized in capturing distinct evolving patterns. The framework independently forecasts each sample of the time series using these models in a decoupled manner. The final prediction is then obtained through a weighted aggregation of the intermediate results. While conceptually similar to ST-ExpertNet \cite{wang2022st}, which employs a mixture-of-experts architecture to capture distinct flow patterns driven by city area functional distributions for traffic prediction, our approach is designed for general time series with potentially diverse evolving patterns, without requiring predefined functional distributions and the corresponding ground truth.

A key challenge in this architecture lies in determining the proper gating function \cite{moe} for a given multivariate time series input $x_i$ so that appropriate model assignments can be obtained for different evolving patterns, especially in the absence of explicit supervision, while ensuring the uniqueness of each forecasting model. To address this challenge, we propose the Forecaster Aware Gate (FAG) module in Section \ref{section:fag}, which adaptively generates routing signals based on the state of each forecasting model and the characteristics of the input series. The forecasters' states are quantified based on the novel Linear Weight Approximation (LWA) strategy, which will be detailed in Section \ref{section:lwa}. Specifically, the routing signals of each input multivariate time series data relative to the backbone models are extracted as follows:
\begin{equation}\label{eq:fag}
\begin{aligned}
    \beta_i &= \operatorname{FAG}(x_i, \Gamma), \\
    \Gamma &= \{\gamma_1, \gamma_2, ..., \gamma_K\},
\end{aligned}
\end{equation}
where $\gamma_j \in \mathbb{R}^{L_{in}\times L_{out}}$ is a compact matrix presenting the transformation function of the $j$-th forecasting model $f_{\theta_j}$, and the learned routing signals $\beta_i \in \mathbb{R}^{C\times K}$ can be regarded as the pattern tendency of $x_i$ towards the backbone forecasters, ensuring that the evolving pattern of a given time series can be adequately addressed by a proper expert.

Consequently, the channel-wise dependence in DisenTS is characterized by a similar evolving pattern, and the future series is predicted by all the experts in the model pool individually in the forecasting stage. The corresponding forecast results $\bar{y}_i^j$ are defined as follows:
\begin{equation}
\bar{y}_i = \sum_{m=1}^K \beta_i^{m} \odot f_{\theta_m}(x_i).
\end{equation}
Here $\odot$ denotes an element-wise product and $\beta_i^{m}\in \mathbb{R}^{C\times 1}$ denotes the pattern tendency vector of all the $C$ channels towards the $m$-th forecasting model. The final forecasting result, $\bar{y}_i$, for the future multivariate time series is obtained through a weighted sum operation. Additionally, to handle the inherent non-stationarity of time series data and produce more robust forecasting results, a series stationary layer \cite{RevIN} is applied within the framework. This layer ensures the proper functioning of the FAG module by mitigating the effects of varying data distributions and scales. In particular, the time series data are normalized using their mean and standard deviation before being input into the framework for forecasting. The output is then denormalized using the same statistical measures to generate the final forecasting results. The normalization and denormalization process is expressed as follows:
\begin{equation}
\begin{aligned}
    x_{i} = \frac{x_i-\mu}{\sigma+\epsilon}\quad &\text{and}\quad \mu = \mathbb{E}[x_i], \sigma^2=\operatorname{Var}[x_i], \\
    \bar{y}_i &= \bar{y}_i \odot (\sigma + \epsilon) + \mu.
\end{aligned}
\end{equation}

\subsection{Forecaster Aware Gate}\label{section:fag}
Although the overall computational process of DisenTS is straightforward, a key challenge arises in generating appropriate routing signals for the input time series data, particularly due to the lack of explicit supervision guiding the optimization of the mixture-of-experts network. To address this, we propose the novel Forecaster Aware Gate (FAG) module, which adaptively generates routing signals by considering both the states of the forecasters and the characteristics of the input series. This approach effectively guides each forecasting model to specialize in distinct channel evolving patterns implicitly.

Specifically, we first project both the input data and the forecasters’ states into a unified embedding space to better capture their interdependencies. For a given input multivariate time series data $x$, a channel-independent projection operation with a learnable matrix $W_{in} \in \mathbb{R}^{L_{in} \times d}$ is applied to map the original channels into a representation space $h_x$, where $d$ denotes the hidden dimension size. Similarly, a multi-layer perceptron (MLP) network is employed to project the transformation matrices, which represent the functions of each forecasting model, into their corresponding hidden representations $h_f$.
\begin{equation}
\begin{aligned}
    h_x &= x \cdot W_{in} \in \mathbb{R}^{C \times d},\\
    h_f &= \operatorname{MLP}(\Gamma) \in \mathbb{R}^{K\times d}.
\end{aligned}
\end{equation}

Based on these representations, a standard transformer block \cite{vaswani2017attention} incorporating dropout, feed-forward layers, and layer normalization is introduced. In this setup, a cross-attention module is employed instead of the conventional self-attention mechanism, where the time series data embeddings serve as the queries, and the hidden representations of the forecasters are used as keys and values. This structure facilitates deep interaction and blending between the time series representation and the forecasters' latent representations. Moreover, by computing the cross-attention matrix, the resulting score map effectively captures the similarity or correlation between the evolving patterns of the time series and the transformation functions of the forecasters. Finally, a prediction layer, parameterized by a learnable matrix $W_{out} \in \mathbb{R}^{d \times K}$, is applied with a softmax activation to generate the forecaster-aware routing signals. The overall computation is formalized as follows:
\begin{equation}
\begin{aligned}
    h_x^{'} &= \operatorname{CrossTrmBlock}(h_x, h_f), \\
    \beta &= \operatorname{Softmax}(h_x^{'} \cdot W_{out}).
\end{aligned}
\end{equation}

\subsection{Linear Weight Approximation}\label{section:lwa}
As discussed in the previous section, DisenTS utilizes $K$ forecasting models to capture the diverse evolving patterns within the data, guided by the FAG module. A natural challenge arises: how can we effectively obtain the representations of these forecasting models, denoted as $\Gamma$, especially when dealing with complex deep neural forecasters, whose parameters can reach millions? To this end, we propose a novel strategy called Linear Weight Approximation (LWA), specifically designed for forecasting tasks. This approach offers precise quantization of complex models, significantly alleviating the challenge.

In particular, for arbitrary given forecasting model $f_\theta$, the model essentially transforms the input time series to generate forecasted results, denoted as $y=f_{\theta}(x)$. This transformation can range from simple statistical methods to complex non-linear deep neural networks. By applying a first-order Taylor expansion around the zero vector $\vec{0}$, the neural transformation function of the forecasting model on the input time series can be approximated as:
\begin{equation}
\begin{aligned}
    y= f_{\theta}(x) &= f_{\theta}(\vec{0}) + x\cdot\nabla f_{\theta}(\vec{0}) + O(||x||^2) \\
    &\approx x\cdot\nabla f_{\theta}(\vec{0}) + \epsilon \\
    &= x\cdot W_{\theta} + \epsilon,
\end{aligned}
\end{equation}
where $W_{\theta}\in \mathbb{R}^{L_{in}\times L_{out}}$ presents the derivative of the model with respect to $x$, and $\epsilon$ accounts for the approximation error between the forecasting results and the approximated ones.

Thus, the forecasting models are simplified and quantified by a linear model derived from a batch of input and prediction pairs. Since these pairs still reside within the original time series space, the linear model's weight $W_{\theta}$ naturally encapsulates the functionality of the forecaster, illustrating how future values are regressed from the observed series via a weighted sum operation in a channel-independent manner. Notably, recent studies \cite{DLinear, sparsetsf} have demonstrated that simple linear transformations are sufficiently effective in capturing most temporal dependencies in time series data, while still achieving competitive performance. This suggests that the approximation error remains within an acceptable range, ensuring that the model's key characteristics are preserved. Furthermore, given that the resulting matrix is relatively small, this method is recognized as a Linear Weight Approximation representation of the forecasting model.

Following this approach, we can estimate the representations of all $K$ backbone models in DisenTS. Rather than directly computing the intricate derivative $\nabla f_{\theta}(\vec{0})$, we opt to calculate the optimal linear weight through regression on batched data: $W_i = \arg\min_W ||x_{\theta_i}\cdot W-f_{\theta_i}(x_{\theta_i})||^2_F$, where $\|\cdot\|_F$ represents the Frobenius norm of a matrix and the computation can be efficiently achieved using numerically stable pseudo-inverse algorithms. Additionally, since our objective is to capture the diverse evolving patterns across multiple channels with distinct forecasting models, we focus on input and prediction pairs that exhibit high correlation with the model for approximation. In practice, we use the $\beta$ values as importance indicators in a top-$k$ sampling operation, selecting the most representative channels and concatenating them as the matched series $\hat{x}_{\theta_i}\in \mathbb{R}^{k \times L_{in}}$ for each backbone model $f_{\theta_i}()$. The corresponding forecasting results are similarly collected, denoted as $f_{\theta_i}(\hat{x}_{\theta_i})$ for brevity. This ensures that the estimated matrix effectively represents the transformation function with minimal error on channels that share similar trends modeled by the backbone. Let $\cdot^+$ denote the pseudo-inverse operation, the linear weight approximation is then formulated as follows:
\begin{equation}\label{eq:lwa}
     W_i = \hat{x}_{\theta_i}^+ \cdot f_{\theta_i}(\hat{x}_{\theta_i}).
\end{equation}

Finally, the framework is optimized using a traditional batched stochastic gradient descent approach. However, shifts in data distribution across batches can result in unstable model representation approximations, ultimately hindering the optimization process. To address this issue, we incorporate the Exponential Moving Average (EMA) paradigm to produce more robust representations, as defined by:
\begin{equation}\label{eq:ema}
    \gamma_i^t = \alpha\gamma_i^{t-1} + (1-\alpha)W_i,
\end{equation}
where $\gamma_i^t$ is the robust representation at iteration $t$, and $\alpha$ is the EMA parameter that controls the rate of weighting decrease. These robust representations are then leveraged to generate adaptive guiding signals in the FAG module through Eq. \ref{eq:fag}.

\subsection{Similarity Constraint Among Forecasters}
Our proposal is to capture multiple channel evolving patterns for \textbf{general} multivariate time series, where no explicit partitions on the patterns are available. This leads to a key challenge: how can we ensure the independence of each backbone model, preventing them from converging into the same representation space? This differs from most existing disentangled learning methods \cite{yao2022temporally, li2022towards}, as our goal here is to \textit{ensure disentanglement between the complex transformation functions of neural networks with extensive parameters}, rather than simply disentangling data representations.

Fortunately, benefiting from the proposed Linear Weight Approximation (LWA) strategy, we can use the approximated compact matrices as reliable representations for the complex functionalities of forecasting networks. Moreover, let $\widetilde{\cdot}$ be the flatten operation, and we can obtain the interpretable vectorized representations $\widetilde{W_i}$. On this basis, we introduce the Similarity Constraint (SC) on these vectors to reduce the similarity between them and therefore ensure implicit disentanglement among backbone forecasting models. In detail, we propose to construct the constraint based on the InfoNCE loss \cite{infonce} which is defined as follows:
\begin{equation}
    L_{SC} = \sum_{i=1}^K -\log \frac{\exp{(\langle \widetilde{W_i}, \widetilde{\gamma_i} \rangle)}}{\sum_{j=1}^K\exp{(\langle \widetilde{W_i}\widetilde{\gamma_j}\rangle)}}
\end{equation}

The constraint operates on two levels. On the one hand, the InfoNCE loss theoretically maximizes the mutual information between positive pairs and minimizes the mutual information between negative pairs \cite{infonce}, forcing each forecaster to specialize in a unique evolving pattern and to avoid mutual interference. Additionally, this constraint encourages uniformity of representations on the hypersphere \cite{wang2020understanding}, which guides the forecasters to uncover deeper, more complex patterns instead of focusing on simple ones, thereby enhancing forecasting performance. On the other hand, we construct sample pairs using the LWA of a given training batch and the corresponding robust approximation obtained through the Exponential Moving Average (EMA). By naturally aligning the representations, this approach reduces the impacts of unstable linear weight approximation during training due to potential data distribution shifts and ensures a stable EMA procedure for the consistent generation of routing signals. This design allows the representations of different backbone models to be appropriately separated in specific directions, helping to ensure effective disentanglement and improve prediction accuracy across diverse input series.

Finally, the proposed DisenTS framework is optimized in a multitask fashion. Let $L_{FC}$ denote the conventional forecasting optimization target of MSE and $\lambda$ presents the weighting parameter, the overall loss function $\mathcal{L}$ over a mini-batch $B$ is defined as:
\begin{equation}
\begin{aligned}
    L_{FC} &= \sum_{i=1}^B\frac{1}{B}||y_i - \bar{y}_i ||_2^2, \\
    \mathcal{L} &= L_{FC} + \lambda L_{SC}.
\end{aligned}
\end{equation}

By jointly optimizing both targets, the DisenTS framework learns to balance both accurate predictions and effective specialization among the backbones, thus enhancing the adaptability of the model to diverse evolving patterns in the multivariate time series data.

\section{Experiments}
In this section, we undertake comprehensive experiments on a well-established benchmark dataset and compare our findings with state-of-the-art methods to clearly demonstrate the efficacy of our proposed DisenTS framework.
\subsection{Experimental Setup}
\subsubsection{Datasets}

We conducted experiments on 14 datasets, covering a wide range of data scopes and forecasting settings. The dataset descriptions are as follows: (1) \textbf{ETT}\footnote{https://github.com/zhouhaoyi/ETDataset} records oil temperature and load metrics from electricity transformers, tracked between July 2016 and July 2018. It is subdivided into four mini-datasets, with data sampled either hourly or every 15 minutes. (2) \textbf{Electricity}\footnote{https://archive.ics.uci.edu/ml/datasets/ElectricityLoadDiagrams20112014} captures the electricity usage of 321 clients, monitored from July 2016 to July 2019. (3) \textbf{Exchange}\footnote{https://github.com/laiguokun/multivariate-time-series-data} contains daily exchange rates of eight nations, spanning from 1990 to 2016. (4) \textbf{Traffic}\footnote{http://pems.dot.ca.gov} provides hourly traffic volume data on San Francisco freeways, recorded by 862 sensors from 2015 to 2016. (5) \textbf{Weather}\footnote{https://www.bgc-jena.mpg.de/wetter/} comprises 21 weather indicators, including air temperature and humidity, collected every 10 minutes in 2021. (6) \textbf{Solar}\footnote{http://www.nrel.gov/grid/solar-power-data.html} dataset records the solar power production of 137 PV plants in 2006, which is sampled every 10 minutes.
(7) \textbf{PEMS}\footnote{https://github.com/guoshnBJTU/ASTGNN/tree/main/data} is another series of traffic flow datasets with four subsets, where traffic information is recorded every 5 minutes by multiple sensors. (8) \textbf{ILI}\footnote{https://gis.cdc.gov/grasp/fluview/fluportaldashboard.html} logs the weekly ratio of patients with influenza-like symptoms to total patients, collected by the U.S. Centers for Disease Control and Prevention from 2002 to 2021. Detailed information about these datasets is provided in Table \ref{tab:dataset}. 
\begin{table}[htbp]
  \centering
  \caption{The Statistics of Each Dataset.}
  \tabcolsep=0.2cm
  \label{tab:dataset}
    \begin{tabular}{ccccc}
    \toprule
    Dataset & Variables & Frequency & Length & Scope \\
    \midrule
    ETTh1\&ETTh2 & 7     & 1 Hour & 17420 & Energy \\
    ETTm1\&ETTm2 & 7     & 15 Minutes & 69680 & Energy \\
    Electricity & 321   & 1 Hour & 26304 & Energy \\
    Exchange & 8     & 1 Day & 7588  & Finacial \\
    Solar & 137   & 10 Minutes & 52560 & Nature \\
    Weather & 21    & 10 Minutes & 52696 & Nature \\
    Traffic & 862   & 1 Hour & 17544 & Transpotation \\
    PEMS03 & 358   & 5 Minutes & 26208 & Transpotation \\
    PEMS04 & 307   & 5 Minutes & 16992 & Transpotation \\
    PEMS07 & 883   & 5 Minutes & 28224 & Transpotation \\
    PEMS08 & 170   & 5 Minutes & 17856 & Transpotation \\
    ILI   & 7     & 1 Week & 966   & Illness \\
    \bottomrule
    \end{tabular}
\end{table}%

\begin{table*}[htbp]
  \small
  \renewcommand\arraystretch{1.0}
  \tabcolsep=0.1cm
  \centering
  \caption{Long-term multivariate time series forecasting results comparing DisenTS with state-of-the-art channel-independent methods. The lookback length is set to 336 for all experimental settings. The \textbf{bold} values indicate a better performance.}
\begin{tabular}{c|c|cccc|cccc|cccc|cccc}
    \toprule
    \multicolumn{2}{c|}{Methods} & \multicolumn{2}{c}{DLinear} & \multicolumn{2}{c}{+ DisenTS} & \multicolumn{2}{c}{SparseTSF} & \multicolumn{2}{c}{+ DisenTS} & \multicolumn{2}{c}{PatchTST} & \multicolumn{2}{c}{+ DisenTS} & \multicolumn{2}{c}{ConvTimeNet} & \multicolumn{2}{c}{+ DisenTS} \\
    \multicolumn{2}{c|}{Metric} & MSE   & MAE   & MSE   & \multicolumn{1}{c}{MAE} & MSE   & MAE   & MSE   & \multicolumn{1}{c}{MAE} & MSE   & MAE   & MSE   & \multicolumn{1}{c}{MAE} & MSE   & MAE   & MSE   & MAE \\
    \midrule
    \multirow{4}[2]{*}{\rotatebox{90}{Electricity}} & 96    & 0.141  & 0.237  & \textbf{0.131 } & \textbf{0.225 } & 0.147  & 0.243  & \textbf{0.137 } & \textbf{0.230 } & 0.138  & 0.233  & \textbf{0.130 } & \textbf{0.224 } & 0.133  & 0.227  & \textbf{0.131 } & \textbf{0.225 } \\
          & 192   & 0.155  & 0.249  & \textbf{0.151 } & \textbf{0.244 } & 0.159  & 0.252  & \textbf{0.153 } & \textbf{0.245 } & 0.153  & 0.247  & \textbf{0.149 } & \textbf{0.242 } & 0.150  & \textbf{0.242 } & \textbf{0.149 } &\textbf{0.242 } \\
          & 336   & 0.172  & 0.265  & \textbf{0.168 } & \textbf{0.261 } & 0.174  & 0.267  & \textbf{0.170 } & \textbf{0.262 } & 0.170  & 0.263  & \textbf{0.166 } & \textbf{0.262 } & 0.166  & 0.259  & \textbf{0.165 } &  \textbf{0.257 } \\
          & 720   & 0.210  & 0.297  & \textbf{0.205 } & \textbf{0.292 } & 0.212  & 0.299  & \textbf{0.209 } & \textbf{0.295 } & 0.206  & 0.295  & \textbf{0.202 } & \textbf{0.293 } & 0.205  & 0.292  & \textbf{0.203 } & \textbf{0.291 } \\
    \midrule
    \multirow{4}[2]{*}{\rotatebox{90}{Exchange}} & 96    & 0.088  & 0.208  & \textbf{0.087 } & \textbf{0.204 } & \textbf{0.088 } & \textbf{0.208 } & 0.089  & 0.209  & 0.094  & 0.216  & \textbf{0.086 } & \textbf{0.207 } & \textbf{0.086 } & \textbf{0.205 } & \textbf{0.086 } & \textbf{0.205 } \\
          & 192   & 0.184  & 0.304  & \textbf{0.182 } & \textbf{0.301 } & \textbf{0.176 } & \textbf{0.299 } & \textbf{0.176 } & \textbf{0.299 } & \textbf{0.191 } & \textbf{0.311 } & 0.194  & 0.317  & \textbf{0.177 } & \textbf{0.298 } & 0.184  & 0.305  \\
          & 336   & \textbf{0.337 } & \textbf{0.421 } & 0.345  & 0.426  & 0.326  & 0.415  & \textbf{0.322 } & \textbf{0.410 } & \textbf{0.343 } & \textbf{0.427 } & 0.350  & 0.431  & \textbf{0.323 } & \textbf{0.411 } & 0.335  & 0.419  \\
          & 720   & 0.885  & 0.704  & \textbf{0.848 } & \textbf{0.695 } & \textbf{0.857 } & 0.694  & 0.860  & \textbf{0.693 } & 0.888  & 0.706  & \textbf{0.847 } & \textbf{0.683 } & \textbf{0.859 } & \textbf{0.692 } & 0.895  & 0.706  \\
    \midrule
    \multirow{4}[2]{*}{\rotatebox{90}{Traffic}} & 96    & 0.411  & 0.280  & \textbf{0.386 } & \textbf{0.269 } & 0.415  & 0.279  & \textbf{0.398 } & \textbf{0.270 } & 0.395  & 0.272  & \textbf{0.365 } & \textbf{0.254 } & 0.381  & 0.264  & \textbf{0.373 } & \textbf{0.258 } \\
          & 192   & 0.423  & 0.285  & \textbf{0.416 } & \textbf{0.282 } & 0.426  & 0.283  & \textbf{0.416 } & \textbf{0.277 } & 0.411  & 0.278  & \textbf{0.390 } & \textbf{0.265 } & 0.399  & 0.271  & \textbf{0.392 } & \textbf{0.266 } \\
          & 336   & 0.437  & 0.291  & \textbf{0.424 } & \textbf{0.286 } & 0.438  & 0.289  & \textbf{0.428 } & \textbf{0.283 } & 0.424  & 0.284  & \textbf{0.407 } & \textbf{0.274 } & 0.410  & 0.276  & \textbf{0.404 } & \textbf{0.272 } \\
          & 720   & 0.463  & 0.306  & \textbf{0.447 } & \textbf{0.299 } & 0.464  & 0.305  & \textbf{0.461 } & \textbf{0.304 } & 0.453  & 0.300  & \textbf{0.438 } & \textbf{0.291 } & 0.440  & 0.294  & \textbf{0.435 } & \textbf{0.290 } \\
    \midrule
    \multirow{4}[2]{*}{\rotatebox{90}{Weather}} & 96    & 0.176  & 0.227  & \textbf{0.158 } & \textbf{0.206 } & 0.177  & 0.227  & \textbf{0.153 } & \textbf{0.200 } & \textbf{0.147 } & \textbf{0.197 } & \textbf{0.147 } & 0.198  & 0.156  & 0.206  & \textbf{0.152 } & \textbf{0.200 } \\
          & 192   & 0.219  & 0.262  & \textbf{0.200 } & \textbf{0.245 } & 0.220  & 0.263  & \textbf{0.198 } & \textbf{0.242 } & 0.191  & \textbf{0.240 } & \textbf{0.190 } & \textbf{0.240 } & 0.201  & 0.248  & \textbf{0.194 } & \textbf{0.240 } \\
          & 336   & 0.267  & 0.297  & \textbf{0.251 } & \textbf{0.283 } & 0.267  & 0.297  & \textbf{0.252 } & \textbf{0.282 } & 0.244  & 0.282  & \textbf{0.242 } & \textbf{0.281 } & 0.250  & 0.285  & \textbf{0.247 } & \textbf{0.280 } \\
          & 720   & 0.335  & 0.344  & \textbf{0.324 } & \textbf{0.335 } & 0.335  & 0.344  & \textbf{0.330 } & \textbf{0.340 } & 0.320  & \textbf{0.334 } & \textbf{0.316 } & \textbf{0.334 } & 0.323  & 0.335  & \textbf{0.321 } & \textbf{0.332 } \\
    \midrule
    \multirow{4}[2]{*}{\rotatebox{90}{ETTh1}} & 24    & 0.371  & \textbf{0.392 } & \textbf{0.370 } & \textbf{0.392 } & 0.394  & 0.407  & \textbf{0.371 } & \textbf{0.392 } & 0.382  & 0.403  & \textbf{0.375 } & \textbf{0.398 } & 0.382  & 0.403  & \textbf{0.374 } & \textbf{0.395 } \\
          & 36    & \textbf{0.404 } & \textbf{0.412 } & \textbf{0.404 } & \textbf{0.412 } & 0.413  & 0.417  & \textbf{0.402 } & \textbf{0.410 } & 0.416  & 0.423  & \textbf{0.412 } & \textbf{0.419 } & 0.416  & 0.423  & \textbf{0.409 } & \textbf{0.417 } \\
          & 48    & 0.431  & 0.430  & \textbf{0.429 } & \textbf{0.429 } & 0.423  & 0.424  & \textbf{0.422 } & \textbf{0.421 } & 0.441  & 0.440  & \textbf{0.437 } & \textbf{0.437 } & 0.439  & 0.437  & \textbf{0.434 } & \textbf{0.436 } \\
          & 60    & 0.450  & 0.462  & \textbf{0.433 } & \textbf{0.451 } & 0.422  & 0.442  & \textbf{0.418 } & \textbf{0.440 } & 0.470  & 0.475  & \textbf{0.447 } & \textbf{0.463 } & 0.438  & 0.456  & \textbf{0.436 } & \textbf{0.454 } \\
    \midrule
    \multirow{4}[2]{*}{\rotatebox{90}{ETTh2}} & 96    & 0.276  & 0.337  & \textbf{0.273 } & \textbf{0.333 } & 0.291  & 0.344  & \textbf{0.287 } & \textbf{0.342 } & \textbf{0.278 } & \textbf{0.341 } & 0.286  & 0.344  & \textbf{0.278 } & \textbf{0.340 } & 0.284  & 0.345  \\
          & 192   & \textbf{0.335 } & \textbf{0.377 } & 0.336  & 0.378  & 0.354  & 0.386  & \textbf{0.344 } & \textbf{0.379 } & \textbf{0.343 } & 0.382  & 0.347  & \textbf{0.381 } & \textbf{0.341 } & 0.381  & \textbf{0.341 } & \textbf{0.380 } \\
          & 336   & \textbf{0.368 } & 0.405  & 0.369  & \textbf{0.403 } & 0.369  & 0.401  & \textbf{0.363 } & \textbf{0.398 } & \textbf{0.372 } & 0.404  & 0.374  & \textbf{0.403 } & \textbf{0.364 } & \textbf{0.403 } & 0.368  & \textbf{0.403 } \\
          & 720   & 0.393  & 0.430  & \textbf{0.391 } & \textbf{0.429 } & 0.393  & 0.426  & \textbf{0.390 } & \textbf{0.425 } & 0.395  & 0.430  & \textbf{0.392 } & \textbf{0.427 } & \textbf{0.393 } & \textbf{0.430 } & 0.398  & 0.431  \\
    \midrule
    \multirow{4}[2]{*}{\rotatebox{90}{ETTm1}} & 96    & 0.304  & 0.347  & \textbf{0.292 } & \textbf{0.341 } & 0.306  & \textbf{0.346 } & \textbf{0.301 } & \textbf{0.346 } & 0.298  & 0.345  & \textbf{0.292 } & \textbf{0.343 } & 0.297  & 0.346  & \textbf{0.291 } & \textbf{0.340 } \\
          & 192   & 0.339  & 0.367  & \textbf{0.330 } & \textbf{0.364 } & 0.340  & \textbf{0.366 } & \textbf{0.334 } & \textbf{0.366 } & 0.339  & 0.374  & \textbf{0.335 } & \textbf{0.372 } & 0.332  & 0.366  & \textbf{0.328 } & \textbf{0.362 } \\
          & 336   & 0.374  & 0.387  & \textbf{0.365 } & \textbf{0.384 } & 0.378  & \textbf{0.388 } & \textbf{0.367 } & 0.389  & 0.381  & 0.401  & \textbf{0.364 } & \textbf{0.390 } & 0.367  & 0.384  & \textbf{0.362 } & \textbf{0.383 } \\
          & 720   & 0.431  & 0.420  & \textbf{0.423 } & \textbf{0.417 } & 0.436  & \textbf{0.420 } & \textbf{0.424 } & 0.422  & 0.428  & 0.431  & \textbf{0.413 } & \textbf{0.422 } & 0.428  & 0.418  & \textbf{0.418 } & \textbf{0.416 } \\
    \midrule
    \multirow{4}[2]{*}{\rotatebox{90}{ETTm2}} & 96    & 0.165  & \textbf{0.253 } & \textbf{0.164 } & \textbf{0.253 } & 0.171  & 0.258  & \textbf{0.168 } & \textbf{0.257 } & 0.174  & 0.261  & \textbf{0.168 } & \textbf{0.255 } & \textbf{0.166 } & 0.255  & \textbf{0.166 } & \textbf{0.251 } \\
          & 192   & 0.220  & \textbf{0.290 } & \textbf{0.219 } & \textbf{0.290 } & \textbf{0.222 } & \textbf{0.292 } & 0.227  & 0.294  & 0.238  & 0.307  & \textbf{0.229 } & \textbf{0.295 } & 0.226  & 0.294  & \textbf{0.222 } & \textbf{0.293 } \\
          & 336   & \textbf{0.274 } & \textbf{0.326 } & \textbf{0.274 } & 0.327  & 0.275  & \textbf{0.327 } & \textbf{0.274 } & \textbf{0.327 } & 0.293  & 0.346  & \textbf{0.283 } & \textbf{0.330 } & 0.283  & 0.331  & \textbf{0.273 } & \textbf{0.327 } \\
          & 720   & 0.368  & \textbf{0.383 } & \textbf{0.366 } & \textbf{0.383 } & 0.372  & \textbf{0.384 } & \textbf{0.367 } & \textbf{0.384 } & 0.373  & 0.401  & \textbf{0.369 } & \textbf{0.385 } & 0.362  & 0.385  & \textbf{0.360 } & \textbf{0.382 } \\
    \midrule
    \multirow{4}[2]{*}{\rotatebox{90}{Solar}} & 96    & 0.226  & 0.257  & \textbf{0.201 } & \textbf{0.233 } & 0.253  & 0.275  & \textbf{0.236 } & \textbf{0.259 } & 0.182  & 0.233  & \textbf{0.173 } & \textbf{0.229 } & 0.197  & 0.241  & \textbf{0.186 } & \textbf{0.235 } \\
          & 192   & 0.257  & 0.273  & \textbf{0.222 } & \textbf{0.251 } & 0.277  & 0.285  & \textbf{0.263 } & \textbf{0.274 } & 0.189  & 0.243  & \textbf{0.187 } & \textbf{0.240 } & 0.214  & 0.255  & \textbf{0.198 } & \textbf{0.244 } \\
          & 336   & 0.280  & 0.285  & \textbf{0.232 } & \textbf{0.259 } & 0.293  & 0.294  & \textbf{0.285 } & \textbf{0.284 } & 0.204  & 0.256  & \textbf{0.196 } & \textbf{0.246 } & 0.222  & 0.263  & \textbf{0.205 } & \textbf{0.250 } \\
          & 720   & 0.279  & 0.285  & \textbf{0.232 } & \textbf{0.262 } & 0.299  & 0.290  & \textbf{0.281 } & \textbf{0.283 } & 0.205  & \textbf{0.258 } & \textbf{0.202 } & 0.260  & 0.221  & 0.264  & \textbf{0.209 } & \textbf{0.225 } \\
    \bottomrule
    \end{tabular}
  \label{tab:main results}%
\end{table*}%

\subsubsection{Backbone Models}
Thanks to our flexible and model-agnostic design, the proposed DisenTS framework can serve as a plugin for various forecasting models. To demonstrate the effectiveness of this framework, we selected mainstream models with diverse architectures and evaluated their performance in multivariate settings. Given that the Linear Weight Approximation operates under the assumption of channel independence, we primarily focus on improving the performance of channel-independent approaches, while also providing generalization evaluations for channel-dependent methods.

Specifically, we chose four representative models as our channel-independent baselines: the linear-model-based DLinear \cite{DLinear} and SparseTSF \cite{sparsetsf}, the transformer-based PatchTST \cite{patchtst}, and the convolution-based ConvTimeNet \cite{cheng2024convtimenet}. These models have been shown to achieve state-of-the-art performance across different architectures and computational complexities.

For the channel-dependent baselines, we included four powerful models with mainstream architectures: the linear-model-based TimeMixer \cite{wang2024timemixer}, the transformer-based iTransformer \cite{liuitransformer} and CARD \cite{wang2024card}, and the convolution-based TimesNet \cite{wu2023timesnet}. These methods utilize various channel-dependence modeling techniques, such as channel-mixing, cross-channel attention, and channel-blending.

\subsubsection{Experiments Details}
We employ ADAM \cite{kingma2014adam} as the default optimizer for all our experiments and evaluate the performance of each model using mean squared error (MSE) and mean absolute error (MAE) metrics. A lower MSE/MAE value indicates superior performance. For DisenTS, the number of backbone forecasting models $K$ is chosen within $[2,10]$ and the weighting parameter $\alpha$ within the range of $[0.001, 0.5]$. In our experiments, we simply instantiate it as a framework composed of $K$ identical predictive models to facilitate a more fair comparison with the baseline, while we argue that it is feasible to construct a DisenTS model with $K$ different backbone forecasters. 

For the implementation of the baseline models, we follow the model configuration and recommended hyper-parameter settings provided in the official code repositories, based on the TSLib framework\footnote{https://github.com/thuml/Time-Series-Library}. Each experiment is repeated three times, and their average metrics are reported. All experiments are implemented using PyTorch \cite{paszke2019pytorch} and are conducted on a single NVIDIA RTX 4090 GPU.

\subsection{Main Results}\label{MR}
We conduct extensive experiments on multivariate time series forecasting across all 14 datasets, which cover a wide range of data scenarios and both long-term and short-term forecasting settings. These experiments demonstrate the effectiveness of the proposed DisenTS framework in comparison to mainstream channel-independent methods. Following standard evaluation protocols, the PEMS and ILI datasets are used to assess short-term forecasting performance. For these datasets, the input series length $L_{in}$ is set to 96, and the target series length $L_{out}$ is chosen as $\{12,24,36,48\}$ for PEMS and $\{24,36,48,60\}$ for ILI. The remaining datasets are evaluated under a long-term forecasting setting, where $L_{in}$ is fixed at 336, and $L_{out}$ is chosen as $\{96,192,336,720\}$ for all models.

We present the long-term forecasting results in Table \ref{tab:main results}. According to the table, we can first observe that all the selected baselines achieve competitive performance on the benchmark datasets, and therefore the residual forecasting error may be significantly dominated by inherent noise or unpredictable distribution shifts in the data. Nonetheless, when enhanced by the DisenTS framework, forecasting accuracy is further improved in most cases. Specifically, the DisenTS-enhanced DLinear improves the MAE in \textbf{28 out of 36} experimental settings, indicating that DisenTS enhances accuracy in approximately \textbf{77\%} of the cases. A similar trend holds for other baseline models, with DisenTS improving prediction accuracy in \textbf{27}, \textbf{29}, and \textbf{28} setups, respectively. Additionally, DisenTS provides substantial relative performance gains for the backbone models. For instance, the DisenTS-enhanced DLinear achieves an average MSE reduction of \textbf{14.7\%} on the Solar dataset and \textbf{7.0\%} on the Weather dataset. Moreover, DisenTS reduces overall MSE by \textbf{2.8\%} and \textbf{2.3\%} for SparseTSF and PatchTST, respectively, averaged across all settings. However, we acknowledge that DisenTS may not always deliver optimal improvements, and in certain tasks, it may offer negligible or even negative effects. We leave the exploration of pre-determining suitable scenarios for DisenTS—potentially using statistics such as Pearson Correlation Coefficient or Cross-Correlation Function—for future research.

\begin{figure*}[htbp]
  \centering
   \includegraphics[width=\linewidth]{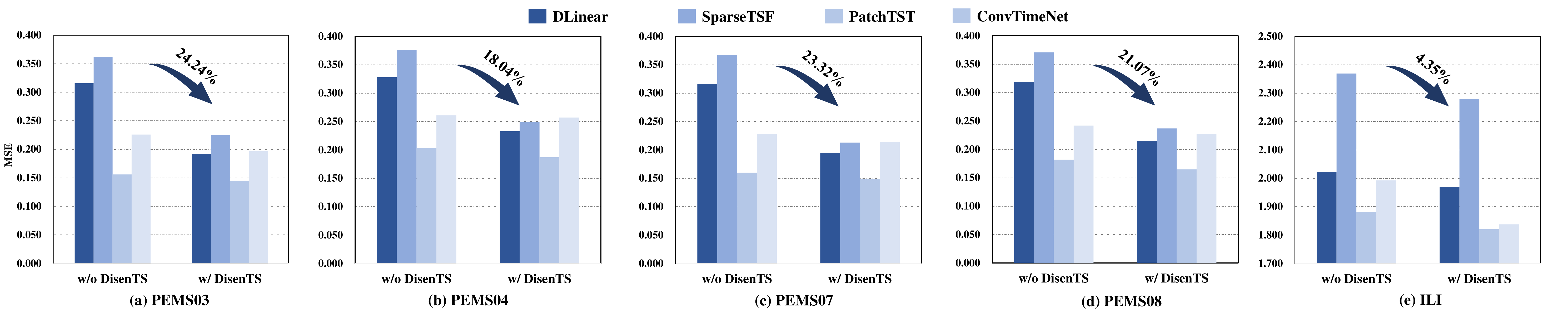}
  \caption{Averaged MSE evaluation on short-term multivariate time series forecasting comparing DisenTS with state-of-the-art channel-independent methods. The lookback length is set to 96 for all experimental settings.}
  \label{fig:short-term-forecasting}
\end{figure*}

On the other hand, we present the MSE evaluations averaged across all experimental settings for each dataset under the short-term forecasting task, as illustrated in Fig. \ref{fig:short-term-forecasting}. The figure also highlights the relative improvements brought by DisenTS, averaged across the baseline models. It is clear that DisenTS achieves superior performance across these datasets and forecasting settings, delivering significant relative improvements. Notably, baseline models enhanced with DisenTS show nearly a \textbf{20\%} average reduction in MSE on the PEMS datasets, further narrowing the performance gap between different model architectures. Given that the PEMS datasets are highly heterogeneous and typically evaluated in a spatio-temporal context, these results strongly validate the effectiveness of DisenTS in handling the diverse evolving patterns of different channels in real-world time series data.

In summary, the proposed DisenTS framework consistently boosts the performance of state-of-the-art channel-independent forecasting models across various forecasting settings and data scenarios. Furthermore, the variation in relative improvements across datasets supports our hypothesis, that real-world multivariate time series data often exhibit distinct evolving patterns, which necessitates tailored disentangled modeling rather than a unified approach.

\subsection{Generalization Study on Channel-Dependent Methods}
As discussed in Section \ref{section:lwa}, a key design of the DisenTS framework involves utilizing a compact linear weight matrix to represent complex forecasting models, estimated through channel-independent regression on input-output pairs. However, for channel-dependent methods, which introduce intricate cross-channel operations, the LWA estimation may not be as suitable, potentially limiting performance. To address this, we aim to explore the generalizability of DisenTS on such approaches.

\begin{table}[htbp]
\tabcolsep=0.065cm
  \centering
  \caption{Performance improvements obtained by DisenTS with channel-dependent forecasting models. The \textbf{bold} values indicate a better performance.}
    \begin{tabular}{c|c|cc|cc|cc|cc}
    \toprule
    \multicolumn{2}{c|}{Methods} & \multicolumn{2}{c}{iTransformer} & \multicolumn{2}{c}{CARD} & \multicolumn{2}{c}{TimesNet} & \multicolumn{2}{c}{TimeMixer} \\
    \multicolumn{2}{c|}{Metric} & MSE   & \multicolumn{1}{c}{MAE} & MSE   & \multicolumn{1}{c}{MAE} & MSE   & \multicolumn{1}{c}{MAE} & MSE   & MAE \\
    \midrule
    \multirow{3}[2]{*}{\rotatebox{90}{Ele.}} & Avg.  & 0.166  & 0.260  & 0.161  & 0.259  & 0.200  & 0.301  & \textbf{0.164 } & \textbf{0.258 } \\
          & w/ DisenTS & \textbf{0.161 } & \textbf{0.256 } & \textbf{0.159 } & \textbf{0.256 } & \textbf{0.194 } & \textbf{0.296 } & \textbf{0.162 } & \textbf{0.255 } \\
          & Imp.(\%) & 3.01  & 1.54  & 1.24  & 1.16  & 3.00  & 1.66  & 1.22  & 1.16  \\
    \midrule
    \multirow{3}[2]{*}{\rotatebox{90}{Traffic}} & Avg.  & 0.388  & 0.276  & 0.396  & 0.280  & 0.624  & 0.336  & 0.422  & 0.287  \\
          & w/ DisenTS & \textbf{0.374 } & \textbf{0.269 } & \textbf{0.394 } & \textbf{0.276 } & \textbf{0.582 } & \textbf{0.318 } & \textbf{0.407 } & \textbf{0.278 } \\
          & Imp.(\%) & 3.61  & 2.54  & 0.51  & 1.43  & 6.73  & 5.36  & 3.55  & 3.14  \\
    \midrule
    \multirow{3}[2]{*}{\rotatebox{90}{Weather}} & Avg.  & 0.237  & 0.273  & 0.232  & 0.270  & 0.249  & 0.286  & 0.229  & 0.269  \\
          & w/ DisenTS & \textbf{0.232 } & \textbf{0.270 } & \textbf{0.229 } & \textbf{0.268 } & \textbf{0.238 } & \textbf{0.277 } & \textbf{0.224 } & \textbf{0.264 } \\
          & Imp.(\%) & 2.11  & 1.10  & 1.29  & 0.74  & 4.42  & 3.15  & 2.18  & 1.86  \\
    \midrule
    \multirow{3}[2]{*}{\rotatebox{90}{Solar}} & Avg.  & 0.230  & 0.270  & 0.199  & 0.262  & 0.241  & 0.279  & 0.236  & 0.275  \\
          & w/ DisenTS & \textbf{0.218 } & \textbf{0.265 } & \textbf{0.193 } & \textbf{0.255 } & \textbf{0.222 } & \textbf{0.268 } & \textbf{0.219 } & \textbf{0.265 } \\
          & Imp.(\%) & 5.22  & 1.85  & 3.02  & 2.67  & 7.88  & 3.94  & 7.20  & 3.64  \\
    \midrule
    \multirow{3}[2]{*}{\rotatebox{90}{ETT}} & Avg.  & 0.375  & 0.402  & 0.358  & 0.389  & 0.399  & 0.421  & 0.372  & 0.396  \\
          & w/ DisenTS & \textbf{0.364 } & \textbf{0.395 } & \textbf{0.351 } & \textbf{0.383 } & \textbf{0.383 } & \textbf{0.409 } & \textbf{0.361 } & \textbf{0.390 } \\
          & Imp.(\%) & 2.93  & 1.74  & 1.96  & 1.54  & 4.01  & 2.85  & 2.96  & 1.52  \\
    \midrule
    \multirow{3}[2]{*}{\rotatebox{90}{PEMS}} & Avg.  & 0.131  & 0.236  & 0.120  & 0.230  & 0.127  & 0.231  & 0.178  & 0.280  \\
          & w/ DisenTS & \textbf{0.124 } & \textbf{0.227 } & \textbf{0.119 } & \textbf{0.229 } & \textbf{0.107 } & \textbf{0.211 } & \textbf{0.146 } & \textbf{0.249 } \\
          & Imp.(\%) & 5.34  & 3.81  & 0.83  & 0.43  & 15.75  & 8.66  & 17.98  & 11.07  \\
    \bottomrule
    \end{tabular}%
  \label{tab:channel-dependent}%
\end{table}%

Specifically, we follow the evaluation protocol outlined in Section \ref{MR} to conduct comparison experiments on both long-term and short-term forecasting using four mainstream channel-dependent methods. The averaged performance across datasets (for ETT and PEMS, results are averaged over 16 experimental setups across four sub-datasets) and the relative improvements enhanced by DisenTS are presented in Table \ref{tab:channel-dependent}. From the table, we can conclude that DisenTS generalizes well on these channel-dependent methods. Although relative improvements in some specific setups may not be immediately apparent, overall, DisenTS consistently delivers positive gains, leading to more precise forecasting results. Moreover, similar trends to those observed with channel-independent models can be observed, where relative improvements on certain datasets (such as Solar and PEMS) are more pronounced. This can be attributed to the characteristics of the data, where DisenTS effectively captures the diverse evolving patterns of heterogeneous datasets, rather than being influenced solely by model architecture. These results further demonstrate the versatility of the DisenTS framework, confirming its role as a model-agnostic plugin that can be seamlessly applied to various forecasting models.

\subsection{Ablation Studies}
The proposed DisenTS framework leverages $K$ distinct forecasting models as experts to implicitly disentangle and capture the diverse evolving patterns across channels. There are two vital components in the framework, the number of experts $K$ and the Similarity Constraint that ensures the disentanglement between experts.
\subsubsection{Number of Experts}
\begin{figure}[htbp]
  \centering
   \includegraphics[width=\linewidth]{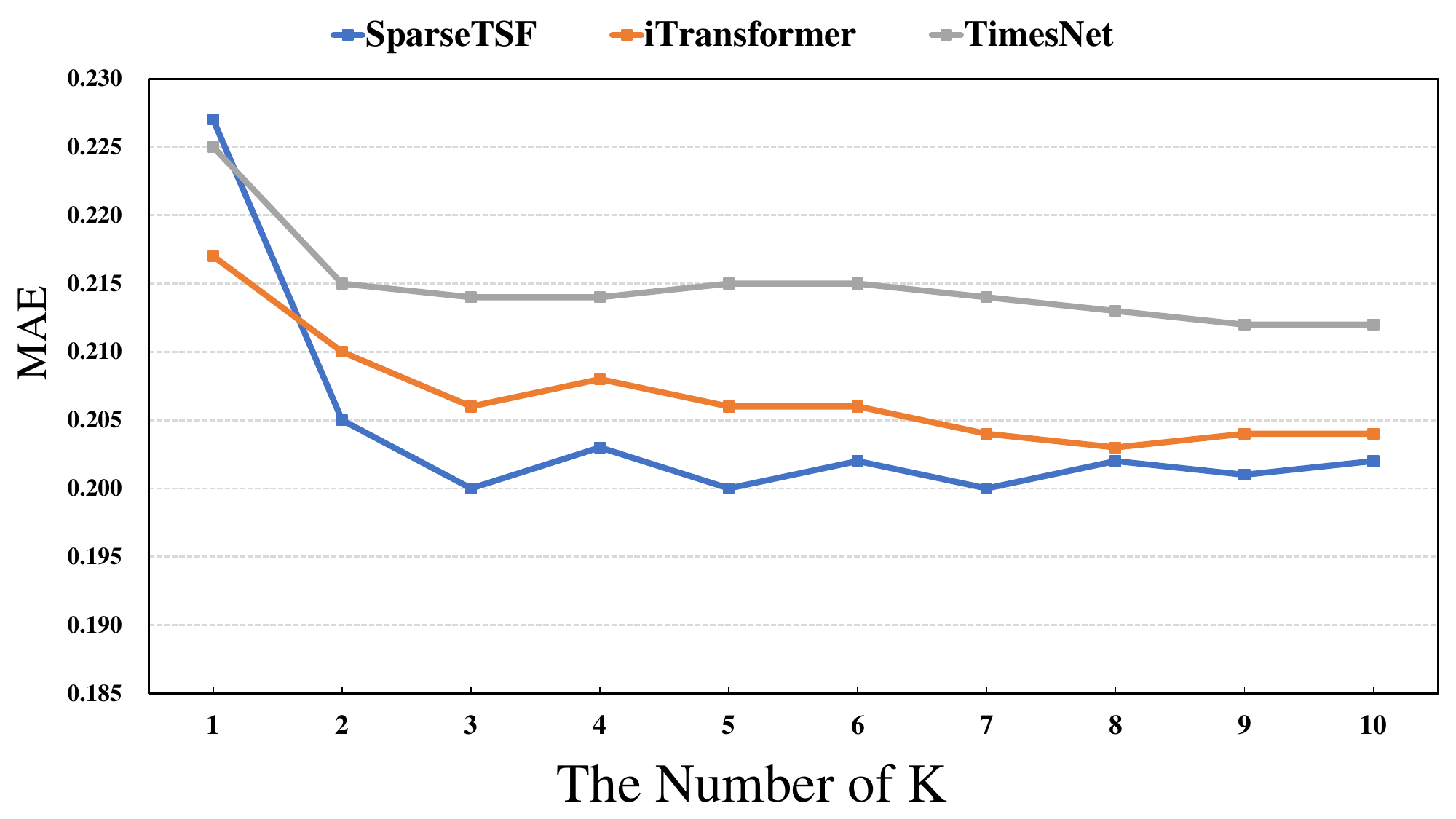}
  \caption{Ablation study on the number of experts. We report the MAE evaluations of DisenTS with different forecasting models and $K$s.}
  \label{fig:SCAnalysis}
\end{figure}

Given the lack of supervision for partitioning different evolving patterns, we manually selected the appropriate value of $K$ based on experimental performance. To further demonstrate the impact of this hyperparameter, we provide the performance results for three baseline models—SparesTSF, iTransformer, and TimesNet—across different architectures with varying values of $K$, as depicted in Fig. \ref{fig:SCAnalysis}.

Specifically, we compare the MAE performance of these models on the weather dataset with a target length of 96. The case where $K$ equals 1 corresponds to the original model without DisenTS enhancement. From the figure, we observe that DisenTS consistently improves forecasting performance across all hyperparameter settings. Even with the least favorable setting, where $K$ is set to 2, DisenTS still achieves relative improvements of \textbf{9.7\%}, \textbf{3.2\%}, and \textbf{4.4\%} for the three models, respectively. Additionally, the results show a clear trend: increasing the number of experts leads to better performance without overfitting, likely due to the implicit disentanglement achieved among the models. Moreover, all three curves exhibit similar marginal effects and optimal parameter settings, further validating that DisenTS is indeed a model-agnostic framework that effectively uncovers the underlying data characteristics, that is, the diverse evolving patterns across channels.

\subsubsection{The Effect of Similarity Constraint}
\begin{table}[htbp]
  \centering
  \caption{Ablation study on the effect of Similarity Constraint. The \textbf{bold} values indicate better performance.}
    
    \begin{tabular}{c|c|cccc}
    \toprule
    \multicolumn{2}{c|}{\multirow{2}[1]{*}{Method}} & \multicolumn{4}{c}{PatchTST} \\
    \multicolumn{2}{c|}{} & \multicolumn{2}{c}{w/ SC} & \multicolumn{2}{c}{w/o SC} \\
    \multicolumn{2}{c|}{Metric} & MSE   & MAE   & MSE   & MAE \\
    \midrule
    \multirow{4}[2]{*}{ETTh1} & 96    & \textbf{0.375 } & \textbf{0.398 } & 0.377  & 0.400  \\
          & 192   & \textbf{0.412 } & \textbf{0.419 } & 0.413  & \textbf{0.419 } \\
          & 336   & \textbf{0.437 } & \textbf{0.437 } & 0.442  & 0.441  \\
          & 720   & \textbf{0.447 } & \textbf{0.463 } & 0.456  & 0.469  \\
    \midrule
    \multirow{4}[2]{*}{Solar} & 96    & \textbf{0.173 } & \textbf{0.229 } & 0.187  & 0.243  \\
          & 192   & \textbf{0.187 } & \textbf{0.240 } & 0.194  & 0.249  \\
          & 336   & \textbf{0.196 } & \textbf{0.246 } & 0.200  & 0.256  \\
          & 720   & \textbf{0.202 } & \textbf{0.260 } & 0.208  & 0.264  \\
    \midrule
    \multirow{4}[2]{*}{PEMS08} & 12    & \textbf{0.089 } & \textbf{0.193 } & 0.092  & 0.198  \\
          & 24    & \textbf{0.137 } & \textbf{0.238 } & 0.145  & 0.250  \\
          & 36    & \textbf{0.191 } & \textbf{0.282 } & 0.204  & 0.297  \\
          & 48    & \textbf{0.241 } & \textbf{0.317 } & 0.265  & 0.340  \\
    \bottomrule
    \end{tabular}%
  \label{tab:ablation}%
\end{table}%

\begin{figure*}[htbp]
  \centering
  \includegraphics[width=\linewidth]{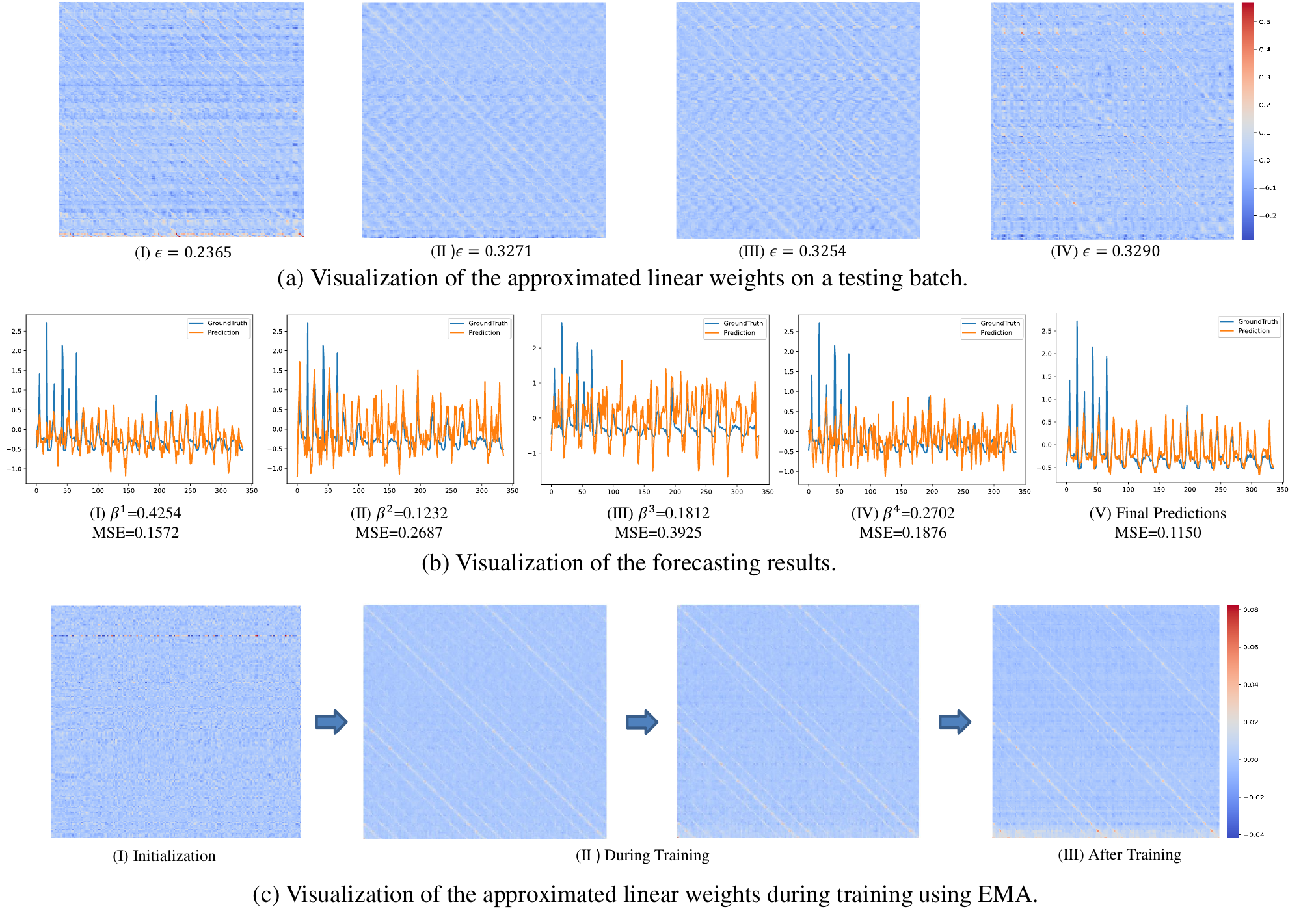}
  \caption{The qualitative visualization of DisenTS-enhanced iTransformer in the Traffic dataset with a prediction length of 336. (a) visualizes the measured transformation matrix of 4 backbones and the corresponding approximation error $\epsilon$ in the first testing batch. (b) presents the forecasting results of all 4 backbones and the final prediction of DisenTS, on a sample of the same testing batch. (c) illustrates how the LWA of the first backbone model changes using EMA in the training procedure.}
  \label{fig:case}
\end{figure*}

The Similarity Constraint (SC) is a key design feature of our framework that ensures the theoretical disentanglement between the forecasting models. In this section, we provide quantitative evaluations of its impact. Specifically, we compare the performance achieved with and without the constraint on the same dataset and backbone model, as shown in Table \ref{tab:ablation}. The results demonstrate that incorporating this constraint consistently improves forecasting accuracy across most experimental settings. In detail, the DisenTS-enhanced PatchTST achieves an average MSE reduction of \textbf{1.1\%} on the ETTh1 dataset, \textbf{3.9\%} on the Solar dataset, and \textbf{6.8\%} on the PEMS08 dataset.

By introducing the explicit pairwise penalty, we push the approximated linear weight representations of the backbones further apart, encouraging greater disentanglement between the models. This enables each model to focus on distinct series patterns, allowing the framework to capture the diverse characteristics of time series data more effectively. As a result, forecasting accuracy improves significantly. These findings underscore the crucial role of the SC in enhancing the overall performance of the DisenTS framework.

\subsection{Qualitative Evaluation}
In this section, we conduct a qualitative analysis to provide deeper insights into the reliability and internal workings of DisenTS, complementing the quantitative performance evaluation. Specifically, we examine the operational principles of the key components within the framework using a DisenTS-enhanced iTransformer model on the Traffic dataset.

As depicted in Fig. \ref{fig:case}(a), we first visualize the approximated linear weights for each backbone model in DisenTS, based on a batch of test series using Eq. \ref{eq:lwa}. Notably, there is a clear separation between the representations, highlighting the successful disentanglement of the learned transformation functions. Additionally, we present the corresponding estimation error across the entire batch of test data, defined as $\epsilon = MSE(f_{\theta_i}(x),xW_i)$, for each backbone in the figure. Despite employing a sampled strategy for computing the linear weights, these representations effectively capture the transformation functions with an acceptable error level across the entire batch of data series, even when using a channel-dependent forecasting model as the backbone. This demonstrates the validity of our approach in quantifying complex forecasters while maintaining reliable performance.

Furthermore, in Fig. \ref{fig:case}(b), we plot the forecasting results for a sampled channel from the testing batch, alongside its evolving pattern tendencies, $\beta$, as computed using Eq. \ref{eq:fag}. This visualization enables us to explore how the framework models the distinct evolving patterns of the input series. As illustrated in the figure, the four backbones produce noticeably different predictions, reaffirming the successful disentanglement among the models. Additionally, the channel demonstrates a clear preference for certain backbone models, as indicated by non-uniform $\beta$ distributions. Delving deeper, we observe that backbone models with higher $\beta$ values generally deliver more accurate future predictions (i.e., lower MSE). This reveals that DisenTS effectively captures the evolving characteristics of each channel series, allowing it to identify the most appropriate model for a given input. The final prediction, which results from a weighted sum of the backbones, outperforms each individual model, highlighting the advantage of this aggregation strategy over a more rigid, hard-sampling approach.

Finally, in Fig. \ref{fig:case}(c), we depict the evolution of the robust Linear Weight Approximation using the Exponential Moving Average (EMA) technique, as outlined in Eq. \ref{eq:ema}. The initialization, which corresponds to the linear weights after the first training batch, appears disorganized. However, due to the stabilizing effect of the EMA strategy, the estimated matrices quickly converge, even during the early training stages. This stabilization allows the FAG module to more effectively leverage the unique strengths of each backbone model, ensuring that it generates reliable guiding signals, even in the presence of distribution shifts among batches. Moreover, explicit patterns emerge in the matrix over time, replacing the initial disarray, and offering the model a degree of interpretability.
\begin{figure*}[htbp]
  \centering
  \includegraphics[width=\linewidth]{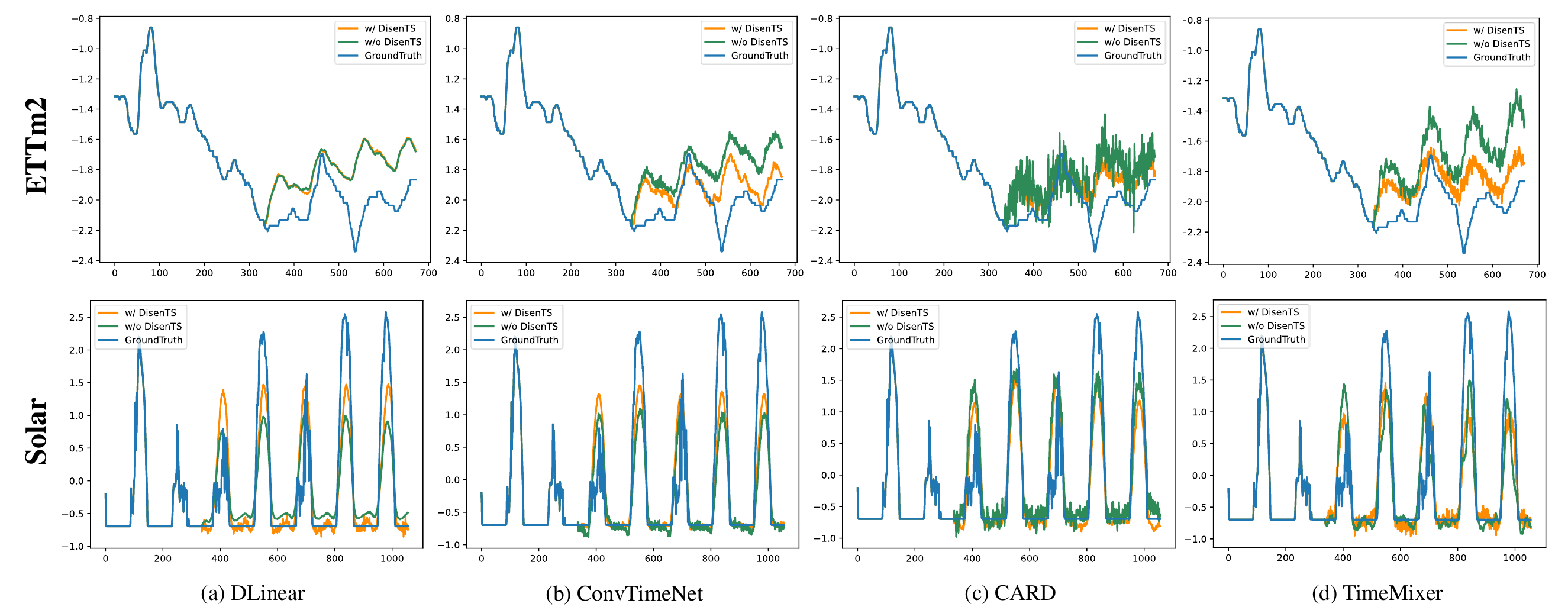}
  \caption{Illustration of long-term forecasting results on a sample of ETTm2 and Solar dataset given by four mainstream forecasting approaches with and without the enhancement of DisenTS.}
  \label{fig:prediction_showcases}
\end{figure*}

\subsection{Prediction Showcases}
In this section, we provide additional prediction showcases to further illustrate the effectiveness of the proposed DisenTS framework, as shown in Fig. \ref{fig:prediction_showcases}. Specifically, we selected DLinear, ConvTimeNet, CARD, and TimeMixer as backbone models, representing both channel-independent and channel-dependent approaches across various architectures. We visualize their forecasting results with and without the enhancement of DisenTS. The results clearly show that DisenTS enables the models to produce more realistic predictions. For instance, on the ETTm2 dataset, the DisenTS-enhanced models deliver predictions that are not only closer to the ground truth but also exhibit greater stability. Similarly, on the Solar dataset, DisenTS improves the backbone model's ability to fit the periodic pattern of invariant values, resulting in more accurate forecasting.

\section{Conclusion}
In this work, we conducted an in-depth study on modeling channel-wise dependence for multivariate time series forecasting. Unlike existing approaches, our objective was to disentangle the diverse evolving patterns present within heterogeneous channels. To achieve this, we introduced DisenTS, a novel, simple yet effective model-agnostic framework. DisenTS leveraged multiple distinct forecasting models in a mixture-of-experts paradigm, with each model tasked with uncovering a unique evolving pattern. To optimize the experts without requiring supervision for partitioning patterns, we proposed a novel FAG module. This module adaptively generated routing signals based on both the forecasters' states and the characteristics of the input series, using the LWA strategy, which quantifies complex networks via compact matrices. Additionally, we incorporated a Similarity Constraint to explicitly ensure disentanglement among the backbone models, guiding each model to specialize in its assigned pattern. Through extensive experiments on a widely-used benchmark dataset, we demonstrated that DisenTS substantially enhances the performance of mainstream forecasting models, achieving state-of-the-art accuracy.

\bibliographystyle{IEEEtran}
\bibliography{sample-base}

\end{document}